\documentclass[11pt,a4paper]{article}
\usepackage[hyperref]{acl2019}
\usepackage{times}
\usepackage{latexsym}

\usepackage{array, caption, floatrow, tabularx, makecell, booktabs}%
\usepackage{hyperref}
\usepackage{color}
\usepackage{colortbl}
\usepackage[colorlinks]{}
\usepackage{multirow}
\usepackage{graphicx}
\usepackage{caption}
\usepackage{subcaption}
\usepackage{url}
\usepackage [autostyle, english = american]{csquotes}
\MakeOuterQuote{"}
\usepackage{fixltx2e}
\usepackage{scalefnt}

\usepackage{algorithm2e}

\usepackage{pgfplots}
\pgfplotsset{compat=1.14}

\makeatletter
\def\BState{\State\hskip-\ALG@thistlm}
\makeatother

\usepackage{url}

\aclfinalcopy 

\title{Curate and Generate: A Corpus and Method for Joint Control of Semantics and Style in Neural NLG}

\author{Shereen Oraby,  Vrindavan Harrison, Abteen Ebrahimi, and Marilyn Walker\\
Natural Language and Dialog Systems Lab \\
University of California, Santa Cruz \\
  {\tt \{soraby,vharriso,aaebrahi,mawalker\}@ucsc.edu} \\}

\date{}

\begin{document}
\maketitle
\begin{abstract}
Neural natural language generation ({\sc nnlg}) from structured
meaning representations has become increasingly popular in recent
years. While we have seen progress with generating syntactically
correct utterances that preserve semantics, various shortcomings of
{\sc nnlg} systems are clear: new tasks require new training data
which is not available or straightforward to acquire, and model
outputs are simple and may be dull and repetitive. This paper addresses 
these two critical challenges in {\sc nnlg} by: (1) scalably (and at no cost) creating training datasets of parallel meaning
representations and reference texts with rich style markup by using
data from freely available and naturally descriptive user reviews, and
(2) systematically exploring how the style markup enables joint 
control of semantic and stylistic aspects of neural model output. We
present {\sc YelpNLG}, a corpus of 300,000 rich, parallel meaning
representations and highly stylistically varied reference texts
spanning different restaurant attributes, and describe a novel
methodology that can be scalably reused to generate {\sc nlg} datasets for
other domains. The experiments show that the models control
important aspects, including lexical choice of
adjectives, output length, and sentiment, allowing the models to
successfully hit multiple style targets without sacrificing semantics.
\end{abstract}

\section{Introduction}
\label{sec:intro}

The increasing popularity of personal assistant dialog systems and the
success of end-to-end neural models on problems such as
machine translation has lead to a surge of interest around data-to-text neural
natural language generation ({\sc nnlg}). State-of-the-art {\sc nnlg}
models commonly use a sequence-to-sequence framework for end-to-end
neural language generation, taking a meaning representation ({\sc mr})
as input, and generating a natural language ({\sc nl}) realization as
output
\cite{DBLP:journals/corr/DusekJ16a,Lampouras2016,Mei2015,Wenetal15}. Table
\ref{table:nlg-datasets} shows some examples of {\sc mr} to human and
system {\sc nl} realizations from recently popular {\sc nnlg}
datasets.

The real power of {\sc nnlg} models over traditional statistical
generators is their ability to produce natural language output from
structured input in a completely data-driven way, without needing
hand-crafted rules or templates. However, these models suffer from two
critical bottlenecks: (1) a {\bf data bottleneck}, i.e. the lack of
large parallel training data of {\sc mr} to {\sc nl}, and (2) a {\bf
  control bottleneck}, i.e. the inability to systematically control
important aspects of the generated output to allow for more
stylistic variation.

\begin{table}[t!]
\begin{footnotesize}
\begin{tabular}
{@{}p{7.8cm}@{}}
\toprule
\bf 1 - E2E \cite{novikova2017e2e} \newline 50k - Crowdsourcing (Domain: Restaurant Description)\\\hline
{\bf MR:} {name[Blue Spice], eatType[restaurant], food[English], area[riverside], familyFriendly[yes], near[Rainbow Vegetarian Cafe]} \newline
{\bf Human:} {\it Situated near the Rainbow Vegetarian Cafe in the riverside area of the city, The Blue Spice restaurant is ideal if you fancy traditional English food whilst out with the kids.} \newline
{\bf System:} Blue Spice is a family friendly English restaurant in the riverside area near Rainbow Vegetarian Cafe.
  \\ \midrule\noalign{\vskip 0.005mm} \toprule 
\bf 2 - WebNLG \cite{nlg_microplanning} \newline 21k - DBPedia and Crowdsourcing (Domain: Wikipedia)  \\\hline
{\bf MR:} { (Buzz-Aldrin, mission, Apollo-11),  (Buzz-Aldrin, birthname, "Edwin Eugene Aldrin Jr."), (Buzz-Aldrin, awards, 20), (Apollo-11, operator, NASA)\newline 
{\bf Human:} \it Buzz Aldrin (born as Edwin Eugene Aldrin Jr) was a crew member for NASA's Apollo 11 and had 20 awards.}  \newline
{\bf System:} Buzz aldrin, who was born in edwin eugene aldrin jr., was a crew member of the nasa operated apollo 11. he was awarded 20 by nasa.
  \\ \midrule\noalign{\vskip 0.005mm} \toprule 


\bf 3 - YelpNLG (this work) \newline 300k - \bf Auto. Extraction (Domain: Restaurant Review)\\\hline
{\bf MR:} (attr=food, val=taco, adj=no-adj, mention=1), (attr=food, val=flour-tortilla, adj=small, mention=1), (attr=food, val=beef, adj=marinated, mention=1), (attr=food, val=sauce, adj=spicy, mention=1)  \newline +[sentiment=positive, len=long, first-person=false, exclamation=false] \newline 
{\bf Human:} {\it The taco was a small flour tortilla topped with marinated grilled beef, asian slaw and a spicy delicious sauce.} \newline
{\bf System:} The taco was a small flour tortilla with marinated beef and a spicy sauce that was a nice touch.
 \\ \bottomrule         
 \end{tabular}
 \caption{A comparison of popular {\sc nnlg} datasets.}
   \label{table:nlg-datasets}
 \end{footnotesize}
\end{table}

Recent efforts to address the data bottleneck with large corpora for
training neural generators have relied almost entirely on high-effort,
costly crowdsourcing, asking humans to write references given an input
{\sc mr}. Table \ref{table:nlg-datasets} shows two recent efforts:
the {\sc e2e nlg} challenge \cite{novikova2017e2e} and the {\sc
  WebNLG} challenge \cite{nlg_microplanning}, both with an example of an
{\sc mr}, human reference, and system realization. The largest dataset, {\sc E2E}, 
consists of 50k instances. Other datasets, such as
the Laptop (13k) and TV (7k) product review datasets, are similar but smaller
\cite{Wen15b,Wenetal15}.

\begin{table}[h!t]
\begin{footnotesize}
\begin{tabular}
{@{}p{7.8cm}@{}}
\toprule
{\bf (1/5 star)} {\it I want to curse everyone I know who recommended this craptacular buffet. [...] It's absurdly overpriced at more than \$50 a 
person for dinner. What do you get for that princely sum? Some
cold crab legs (it's NOT King Crab, either, despite what others are saying) Shrimp cocktail (several of which weren't even deveined. GROSS. [...])} \\ \hline
{\bf (5/5 star)} {\it One of my new fave buffets in Vegas! Very cute interior, and lots of yummy foods! [...] The delicious Fresh, delicious king grab legs!! [...]REALLY yummy desserts! [...] All were grrreat, but that tres leches was ridiculously delicious.}\\ 
\bottomrule         
 \end{tabular}
\vspace{-.1in}
 \caption{Yelp restaurant reviews for the same business.}
   \label{table:yelp}
 \end{footnotesize}
\end{table}

These datasets were created primarily to focus on the task of semantic fidelity, and thus it is very evident from comparing the human and system outputs from each system that the model realizations are less fluent, descriptive, and natural than the human reference. Also, the nature of the domains (restaurant description, Wikipedia infoboxes, and technical product reviews) are not particularly descriptive, exhibiting little variation.

Other work has also focused on the control bottleneck in {\sc nnlg}, but has zoned in on one particular dimension of style, such as sentiment, length, or formality \cite{fancontrollable,hu2017toward,Ficler17,shen2017style,herzig2017neural,fu2018style,RaoTetreault18}. However, human language actually involves a constellation of interacting aspects of style, and {\sc nnlg} models should be able to jointly control these multiple interacting aspects. 

In this work, we tackle {\bf both} bottlenecks simultaneously by
leveraging masses of freely available, highly descriptive user review
data, such as that shown in Table \ref{table:yelp}. These
naturally-occurring examples show a highly positive and highly
negative review for the same restaurant, with many examples of rich
language and detailed descriptions, such as {\it "absurdly
  overpriced"}, and {\it "ridiculously delicious"}. Given the richness
of this type of free, abundant data, we ask: (1) can this freely
available data be used for training {\sc nnlg} models?, and (2) is it
possible to exploit the variation in the data to develop models that jointly control
multiple interacting aspects of semantics and style?

We address these questions by creating the {\sc YelpNLG} corpus,
consisting of 300k {\sc mr} to reference pairs for training {\sc
  nnlg}s, collected {\it completely automatically} using freely
available data (such as that in Table \ref{table:yelp}), and
off-the-shelf tools.\footnote{\url{https://nlds.soe.ucsc.edu/yelpnlg}} Rather than starting with a meaning representation and collecting human references, we begin with the
references (in the form of review sentences), and work backwards --
systematically constructing meaning representations for the sentences
using dependency parses and rich sets of lexical, syntactic, and
sentiment information, including ontological knowledge from
DBPedia. This method uniquely exploits existing
data which is naturally rich in semantic content, emotion, and varied
language. Row 3 of Table \ref{table:nlg-datasets} shows an example {\sc
  mr} from {\sc YelpNLG}, consisting of relational tuples of attributes,
values, adjectives, and order information, as well as sentence-level
information including sentiment, length, and pronouns.

Once we have created the {\sc YelpNLG} corpus, we are in the unique
position of being able to explore, for the first time, how varying
levels of supervision in the encoding of content, lexical choice,
and sentiment can be exploited to control style in {\sc nnlg}. Our contributions
include:\vspace{-.1in}
\begin{itemize}\setlength\itemsep{-0.1em}
\item A new corpus, {\sc YelpNLG}, larger and more lexically and stylistically varied than existing {\sc nlg} datasets;
\vspace{-.05in}
\item A method for creating corpora such as {\sc YelpNLG}, which should be applicable to other domains;
\vspace{-.05in}
\item Experiments on controlling multiple interacting aspects of style 
with an {\sc nnlg}  while maintaining semantic fidelity, and results using a broad range of 
evaluation methods;
\vspace{-.05in}
\item The first experiments, to our knowledge, showing that an {\sc nnlg} can
be trained to control lexical choice of adjectives.
\end{itemize}

We leave a detailed review of prior work
to Section \ref{sec:related} where we can compare it with our own.

\begin{figure*}[t!h]
  \centering
    \includegraphics[width=0.9\textwidth]{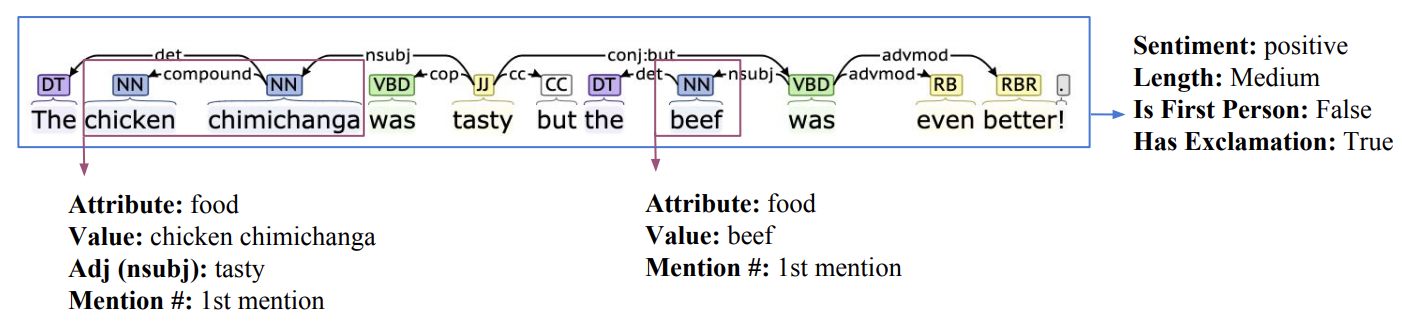}
\caption{Extracting information from a review sentence parse to create an {\sc mr}.}
\vspace{-.1in}
\label{fig:extraction-method}
\end{figure*}

\section{Creating the YelpNLG Corpus}
\label{sec:data}

We begin with reviews from the Yelp challenge dataset,\footnote{{\url{https://www.yelp.com/dataset/challenge}}} which is publicly available and includes structured information for attributes such as location, ambience, and parking availability for over 150k businesses, with around 4 million reviews in total. We note that this domain and dataset are particularly unique in how naturally descriptive the language used is, as exemplified in Table \ref{table:yelp}, especially compared to other datasets previously used for {\sc nlg} in domains such as Wikipedia.

For corpus creation, we must first sample sentences from
reviews in such a way as to allow the automatic and reliable 
construction of {\sc mr}s using fully automatic tools. To identify restaurant
attributes, we use restaurant lexicons from our previous work on template-based {\sc NLG} \cite{oraby2017styleyelp}. The lexicons include five attribute types prevalent in restaurant
reviews: {\it restaurant-type, cuisine, food, service,} and {\it
  staff} collected from Wikipedia and DBpedia, including, for example, around
4k for {\it foods} (e.g. "sushi"), and around 40 for {\it cuisines}
(e.g. "Italian"). 
We then expand these basic 
lexicons by adding in attributes for {\it ambiance} (e.g. "decoration") and {\it price} (e.g. "cost")
using vocabulary items from the E2E generation challenge \cite{Novikovaetal17}.

\begin{table*}[t!]
\begin{small}
\begin{tabular}
{@{} p{0.001in}p{6in} @{}} \toprule
1 & {\bf The chicken chimichanga was tasty but the beef was even better!}   \\
& {\it (attr=}food,  {\it val=}chicken\_chimichanga,  {\it adj=}tasty,  {\it mention=}1), ({\it attr=}food,  {\it val=}beef,  {\it adj=}no\_adj,  {\it mention=}1) \newline +[{\it sentiment=}positive,  {\it len=}medium,  {\it first\_person}=false,  {\it exclamation=}true] \\ \midrule
2 & {\bf Food was pretty good ( i had a chicken wrap ) but service was crazy slow.}   \\
& {\it (attr=}food,  {\it val=}chicken\_wrap,  {\it adj=}no\_adj,  {\it mention=}1), ({\it attr=}service,  {\it val=}service,  {\it adj=}slow,  {\it mention=}1)  \newline +[{\it sentiment=}neutral,  {\it len=}medium,  {\it first\_person=}true,  {\it exclamation=}false] \\ \midrule
3 & {\bf The chicken was a bit bland ; i prefer spicy chicken or well seasoned chicken.}   \\
& {\it (attr=}food,  {\it val=}chicken,  {\it adj=}bland,  {\it mention=}1), ({\it attr=}food,  {\it val=}chicken,  {\it adj=}spicy,  {\it mention=}2), ({\it attr=}food,  {\it val=}chicken,  {\it adj=}seasoned,  {\it mention=}3) +[{\it sentiment=}neutral,  {\it len=}medium,  {\it first\_person=}true,  {\it exclamation=}false]\\ \midrule
4 & {\bf The beef and chicken kebabs were succulent and worked well with buttered rice, broiled tomatoes and raw onions.}   \\
& ({\it attr=}food,  {\it val=}beef\_chicken\_kebabs,  {\it adj=}succulent,  {\it mention=}1), ({\it attr=}food,  {\it val=}rice,  {\it adj=}buttered,  {\it mention=}1), \newline ( {\it attr=}food,  {\it val=}tomatoes,  {\it adj=}broiled,  {\it mention=}1), ({\it attr=}food,  {\it val=}onions,  {\it adj=}raw,  {\it mention=}1)  \newline +[{\it sentiment=}positive,  {\it len=}long,  {\it first\_person=}false,  {\it exclamation=}false]
\\ \bottomrule
\end{tabular}
\vspace{-.1in}
 \caption{Sample sentences and automatically generated {\sc mr}s from {\sc YelpNLG}\label{table:corpora}. Note the stylistic variation that is marked up in the {\sc +style MR}s, especially compared to those in other corpora such as {\sc E2E} or {\sc WebNLG}.}
\end{small}
\end{table*}

To enforce some semantic constraints and "truth grounding" when
selecting sentences without severely limiting variability, we only select sentences that
mention particular food values. A pilot analysis of random reviews
show that some of the most commonly mentioned foods are meat items,
i.e. {\it "meat", "beef", "chicken", "crab",} and {\it
  "steak"}. Beginning with the original set of over 4 million business
reviews, we sentence-tokenize them and randomly sample a set of
500,000 sentences from restaurant reviews that mention of at least one
of the meat items (spanning around 3k unique restaurants, 170k users, and
340k reviews).

We filter to select sentences that are between 4 and 30 words in length: restricting the length increases the likelihood of a successful parse and reduces noise in the process of automatic {\sc mr} construction. We parse the sentences using Stanford dependency parser \cite{stanfordparser}, removing any sentence that is tagged as a fragment. We show a sample sentence parse in Figure \ref{fig:extraction-method}. We identify all nouns and search for them in the attribute lexicons, constructing {\it (attribute, value)} tuples if a noun is found in a lexicon, including the full noun compound if applicable, e.g. {\it (food, chicken-chimichanga)} in Figure \ref{fig:extraction-method}.\footnote{Including noun compounds allows us to identify new values that did not exist in our lexicons, thus automatically expanding them.}  Next, for each {\it (attribute, value)} tuple, we extract all {\it amod}, {\it nsubj}, or {\it compound} relations between a noun value in the lexicons and an adjective using the dependency parse, resulting in {\it (attribute, value, adjective)} tuples. We add in {\it "mention order"} into the tuple distinguish values mentioned multiple times in the same reference.

We also collect sentence-level information {to encode additional style variables}. For {\it sentiment}, we tag each sentence with the sentiment inherited from the "star rating"
of the original review it appears in, binned into one of three values
for lower granularity: {\bf 1} for low review scores (1-2 stars), {\bf
  2} for neutral scores (3 star), and {\bf 3} for high scores (4-5
stars).\footnote{A pilot experiment comparing this method with 
  Stanford sentiment \cite{stanfordsentiment} showed
that copying down the original review ratings gives more
  reliable sentiment scores.}  To experiment with control
of length, we assign a {\it length} bin of {\it
  short} ($\le$ 10 words), {\it medium} (10-20 words), and {\it long}
($\ge$ 20 words). We also include whether the sentence is in first person.


For each sentence, we create 4 {\sc mr} variations.
The simplest variation, {\sc base}, contains only attributes
and their values. The {\sc +adj} version adds adjectives, {\sc +sent}
adds sentiment, and finally the richest {\sc mr}, {\sc +style}, adds
style information on mention order, whether the sentence is first
person, and whether it contains an exclamation. 
Half of the sentences are in first person and around
10\% contain an exclamation, and both of these can contribute to
controllable generation: previous work has explored the effect of
first person sentences on user perceptions of dialog systems \cite{BoyceGorin96}, and
exclamations may be correlated with aspects of a hyperbolic style.

Table \ref{table:corpora} shows sample sentences for the richest version of the {\sc mr} ({\sc +style}) that we create. In Row 1, we see the {\sc mr} from the example in Figure \ref{fig:extraction-method}, showing an example of a {\sc NN} compound, "chicken chimichanga", with adjective "tasty", and the other food item, "beef", with no retrieved adjective. Row 2 shows an example of a "service" attribute with adjective "slow", in the first person, and neutral sentiment. Note that in this example, the method does not retrieve that the "chicken wrap" is actually described as "good", based on the information available in the parse, but that much of the other information in the sentence is accurately captured. We expect the language model to successfully smooth noise in the training data caused by parser or extraction errors.\footnote{{We note that the Stanford dependency parser \cite{stanfordparser} has a token-wise labeled attachment score (LAS) of 90.7, but point out that for our MRs we are primarily concerned with capturing {\sc NN} compounds and adjective-noun relations, which we evaluate in Section \ref{sec:corpus-quality}.}} Row 3 shows an example of the value "chicken" mentioned 3 times, each with different adjectives ("bland", "spicy", and "seasoned"). Row 4 shows an example of 4 foods and very positive sentiment.

{\subsection{Comparison to Previous Datasets}}
Table~\ref{table:compare-corpora} compares {\sc YelpNLG} to previous
work in terms of data size, unique vocab and adjectives,
entropy,\footnote{We show the formula for entropy in Sec \ref{sec:eval}
  on evaluation.} average reference length (RefLen), and examples of stylistic and structural variation in
terms of contrast (markers such as "but" and "although"),
and aggregation (e.g. "both" and "also") \cite{juraskainlg}, showing how our
dataset is much larger and more varied than previous work. We note that the Laptop and E2E datasets (which allow multiple sentences per references) have longer references on average than YelpNLG (where references are always single sentences and have a maximum of 30 words). We are interested in experimenting with longer references, possibly with multiple sentences, in future work.

\begin{figure}[h!]
\begin{subfigure}[b]{0.9\columnwidth}
\centering
        \resizebox {\columnwidth} {!} {
\begin{tikzpicture}
        \begin{axis}[
        scaled ticks=false, 
        tick label style={/pgf/number format/fixed},
ymin=1,
ymax=14000,
legend style={at={(0,1)},anchor=north west,draw=none},
  xticklabel style = {rotate=30,anchor=east},
   enlargelimits = false,
   ylabel=Number of MRs,
   xlabel=Number of Attributes per MR,
   yticklabels={,,2000,4000,6000,8000,10000,12000,14000},
  xticklabels from table={mr-dist-train.dat}{Len},xtick=data]
\addplot[smooth,blue,thick,mark=square*] table [y=Freq,x=X]{mr-dist-train.dat};
\end{axis}
\end{tikzpicture}
}
\end{subfigure}
\caption{MR distribution in {\sc YelpNLG} train.}
    \label{fig:mr-dist}
\end{figure}
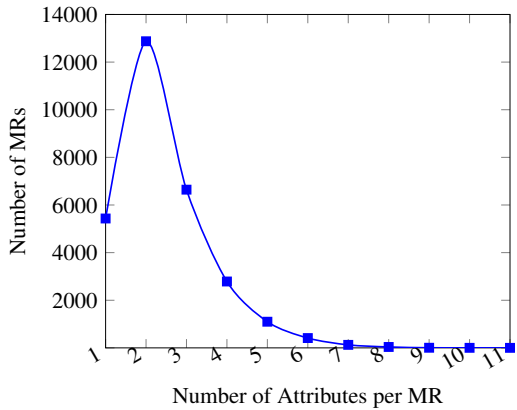

{Figure~\ref{fig:mr-dist} shows the distribution of {\sc mr} length, in terms of the number of attribute-value tuples. There is naturally a higher density of shorter {\sc mr}s, with around 13k instances from the dataset containing around 2.5 attribute-value tuples, but that the {\sc mr}s go up to 11 tuples in length.}

\begin{table}[h!]
\begin{footnotesize}
\begin{tabular}{p{2.7cm}p{1cm}p{1cm}p{1.2cm}}\toprule
 &  {\sc E2E}          & \sc Laptop    & \sc YelpNLG    \\ \hline
Train Size & 42k &	8k & \bf 235k \\
Train Vocab & 2,786 & 1,744 &\bf 41,337 \\
Train \# Adjs & 944 & 381 & \bf 13,097 \\ 
Train Entropy & 11.59 & 11.57 &\bf 15.25 \\ 
Train RefLen &22.4 & \bf 26.4 & 17.32 \\ \midrule
\% Refs w/ Contrast & 5.78\% & 3.61\% &\bf 9.11\% \\
\% Refs w/ Aggreg. & 1.64\% & 2.54\% &\bf 6.39\% \\
\bottomrule
\end{tabular}
\end{footnotesize}
\vspace{-.1in}
\centering \caption{\label{table:compare-corpora} {NLG corpus statistics from {\sc E2E} \cite{novikova2017e2e}, {\sc Laptop} \cite{Wen2016MultidomainNN}, and {\sc YelpNLG} (this work).}}
\end{table}

{ \subsection{Quality Evaluation}
\label{sec:corpus-quality}
We examine the quality of the {\sc mr} extraction with a qualitative study evaluating {\sc YelpNLG} {\sc mr} to {\sc nl} pairs on various dimensions. Specifically, we evaluate {\bf content preservation} (how much of the {\sc mr} content appears in the {\sc nl}, {specifically, nouns and their corresponding adjectives from our parses}), {\bf fluency} (how "natural sounding" the {\sc nl} is, aiming for both grammatical errors and general fluency), and {\bf sentiment} (what the perceived sentiment of the {\sc nl} is). We note that we conduct the same study over our {\sc nnlg} test outputs when we generate data using {\sc YelpNLG} in Section \ref{human-eval}.

We randomly sample 200 {\sc mr}s from the {\sc YelpNLG} dataset, along with their corresponding {\sc nl} references, and ask 5 annotators on Mechanical Turk to rate each output on a 5 point Likert scale (where 1 is low and 5 is high for content and fluency, and where 1 is negative and 5 is positive for sentiment). For content and fluency, we compute the average score across all 5 raters for each item, and average those scores to get a final rating for each model, such that higher content and fluency scores are better. We compute sentiment error by {converting the judgments into 3 bins to match the Yelp review scores (as we did during {\sc mr} creation), finding the average rating for all 5 annotators per item, then computing the {\it difference} between their average score and the true sentiment rating in the reference text (from the original review), such that lower sentiment error is better.

The average ratings for content and fluency are high, at 4.63 and 4.44 out of 5, respectively, meaning that there are few mistakes in marking attribute and value pairs in the {\sc nl} references, and that the references are also fluent. This is an important check because correct grammar/spelling/punctuation is not a restriction in Yelp reviews. For sentiment, the largest error is 0.58 {(out of 3)}, meaning that the perceived sentiment by raters does not diverge greatly, on average, from the Yelp review sentiment assigned in the {\sc mr}, and indicates that inheriting sentence sentiment from the review is a reasonable heuristic.} 

\section{Model Design}
\label{sec:model}

In the standard RNN encoder-decoder architecture commonly used for machine translation \cite{sutskever2014sequence,DBLP:journals/corr/BahdanauCB14}, the probability of a target sentence $w_{1:T}$ given a source sentence $x_{1:S}$ is modeled as $p(w_{1:T}|x) = \prod_{1}^{T} p(w_{t} | w_{1:t-1}, x)$ \cite{OpenNMT}. 

In our case, the input is not a natural language source sentence as in traditional machine translation; instead, the input $x_{1:S}$ is a meaning representation, where each token $x_n$ is itself a tuple of attribute and value features, $(f_{attr}, f_{val})$. Thus, we represent a given input $x_{1:S}$ as a sequence of attribute-value pairs from an input {\sc mr}. For example, in the case of {\sc base MR} {\it[(attr=food, val=steak), (attr=food, val=chicken)]}, we would have $x=x_{1},x_{2}$, where {\it $x_1$=($f_{attr}$=food,$f_{val}$=steak)}, and {\it $x_2$=($f_{attr}$=food,$f_{val}$=chicken)}. The target sequence is a natural language sentence, which in this example might be, {\it "The steak was extra juicy and the chicken was delicious!"}\\

\noindent{\bf Base encoding.} During the encoding phase for {\sc base MR}s, the model takes as input the {\sc mr} as a sequence of attribute-value pairs. We pre-compute separate vocabularies for attributes and values. {\sc mr} attributes are represented as vectors and {\sc mr} values are represented with reduced dimensional embeddings that get updated during training. The attributes and values of the input {\sc mr} are concatenated to produce a sequence of attribute-value pairs that then is encoded using a multi-layer bidirectional LSTM \cite{hochreiter1997long}.\\

\noindent{\bf Additional feature encoding.}
For the {\sc +adj, +sent,} and {\sc +style MR}s, each {\sc mr} is a longer relational tuple, with additional style feature information to encode, such that an input sequence $x_{1:S}=(f_{attr}, f_{val}, f_{1:N})$, and where each $f_n$ is an additional feature, such as adjective or mention order. Specifically in the case of {\sc +style MR}s, the additional features may be sentence-level features, such as sentiment, length, or exclamation.

In this case, we enforce additional constraints on the models for {\sc
  +adj, +sent}, and {\sc +style}, changing the conditional probability
computation for $w_{1:T}$ given a source sentence $x_{1:S}$ to
$p(w_{1:T}|x) = \prod_{1}^{T} p(w_{t} | w_{1:t-1}, x, f)$, where $f$
is the set of new feature constraints to the model.

We represent these additional features as a vector of additional
supervision tokens or side constraints \cite{sennrich16}.
Thus, we construct a vector for each set
of features, and concatenate them to the end of each attribute-value
pair, encoding the full sequence as for {\sc base} above.\\

\noindent{\bf Target decoding.}  At each time step of the decoding
phase the decoder computes a new decoder hidden state based on the
previously predicted word and an attentionally-weighted average of the
encoder hidden states. The conditional next-word distribution $p(w_{t}
| w_{1:t-1}, x, f)$ depends on {\it f}, the stylistic feature constraints
added as supervision. This is produced using the decoder hidden state to
compute a distribution over the vocabulary of target side words. The
decoder is a unidirectional multi-layer LSTM and attention is
calculated as in \citet{luong2015effective} using the \textit{general}
method of computing attention scores. We present model configurations 
in Appendix \ref{sec:app-model}.

\section{Evaluation}
\label{sec:eval}
To evaluate whether the models effectively hit semantic and stylistic targets, 
we randomly split the {\sc YelpNLG} corpus into
80\% train ($\sim$235k instances), 10\% dev and test ($\sim$30k instances each), 
and create 4 versions of the corpus: {\sc base, +adj, +sent,} and {\sc +style}, each with the same split.\footnote{{Since we randomly split the data, we compute the overlap between train and test for each corpus version, noting that around 14\% of test {\sc mr}s exist in training for the most specific {\sc +style} version (around 4.3k of the 30k), but that less than 0.5\% of the 30k full {\sc mr}-ref pairs from test exist in train.}}

Table \ref{table:output} shows examples of output generated by the
models for a given test {\sc mr}, showing the effects of training
models with increasing information. Note that we present the longest
version of the {\sc mr} (that used for the {\sc +style} model), so the
{\sc base}, {\sc +adj}, and {\sc +sent} models use the same {\sc mr}
minus the additional information. Row 1 shows an example of partially correct sentiment for 
{\sc base}, and fully correct sentiment for the
 rest; {\sc +adj} gets the adjectives right, {\sc +sent} is more
 descriptive, and {\sc +style} hits all targets. Row 2 gives an
 example of extra length in {\sc +style}, {\it "the meat was so tender
   and juicy that it melted in your mouth"}. Row 3 shows an
 example of a negative sentiment target, which is achieved by both the
 {\sc +sent} and {\sc +style} models, with interesting descriptions
 such as {\it "the breakfast pizza was a joke"}, and {\it "the pizza
   crust was a little on the bland side"}. 
   We show more {\sc +style} model outputs in Appendix \ref{sec:app-outputs}.

\begin{table*}[t!]
\begin{small}
\begin{tabular}
{@{} p{0.001in}p{.4in}p{5.5in} @{}} \toprule
1 & {\sc mr} &\bf  (food, porridge, no\_adj, mention=1), (food, meat, no\_adj, mention=1), (food, flavor, rich, mention=1), (food, soup, no\_adj, mention=1)  +[sentiment=positive, len=long, first\_person=false, exclamation=false] \\\hline
 & {\sc Base} &  The porridge was good, but the meat lacked flavor and the soup was bland. \\
 & {\sc  +Adj} &  The porridge had a lot of meat in it and the flavor of the soup was rich. \\
 & {\sc  +Sent} &  The porridge had a lot of meat in it and the flavor of the soup was rich and delicious.  \\
 & {\sc  +Style }&  The porridge had a good amount of meat and rich flavor, and the soup was cooked perfectly.  \\ \midrule\noalign{\vskip 0.5mm} \midrule 

 
2 & {\sc mr} &\bf  (food, gyro\_salad, no\_adj, mention=1), (food, meat, no\_adj, mention=1) \newline  +[sentiment=positive, len=long, first\_person=true, exclamation=false] \\\hline
 & {\sc Base} &  I had the gyro salad and the meat was very good. \\
 & {\sc  +Adj} &  I had the gyro salad and the meat was tender and juicy. \\
 & {\sc  +Sent} &  I had the gyro salad and the meat was tender.  \\
 & {\sc  +Style }&  I had the gyro salad and the meat was so tender and juicy that it melted in your mouth.  \\ \midrule\noalign{\vskip 0.5mm} \midrule 
 
3 & {\sc mr} &\bf  (food, eggs, no\_adj, mention=1), (food, ham\_steak, small, mention=1), (food, bacon, chewy, mention=1), (food, breakfast\_pizza, no\_adj, mention=1)  \newline +[sentiment=negative, len=long, first\_person=true, exclamation=false] \\\hline
 & {\sc Base} &  I had the eggs, ham steak, bacon, and buffalo pizza. \\
 & {\sc  +Adj} &  Eggs, ham steak, chewy bacon, and breakfast pizza. \\
 & {\sc  +Sent} &  The eggs were over cooked, the ham steak was small, the bacon was chewy, and the breakfast pizza was a joke.  \\
 & {\sc  +Style }&  I ordered the eggs benedict and the ham steak was small, the bacon was chewy and the pizza crust was a little on the bland side.
\\ \bottomrule
\end{tabular}
 \caption{Sample test {\sc mr} and corresponding outputs for each model. \label{table:output}
Note that the {\sc mr} presented is for {\sc +style}: the other models all provide less information
as described in Section~\ref{sec:data}.}
\end{small}
\end{table*}

\subsection{Automatic Semantic Evaluation}

\noindent\textbf{Machine Translation Metrics.} We begin with an automatic
evaluation using standard metrics frequently used for machine
translation. We use the script provided by the E2E Generation
Challenge\footnote{\url{https://github.com/tuetschek/e2e-metrics}} to
compute scores for each of the 4 model test outputs compared to the
original Yelp review sentences in the corresponding test set. Rows 1-4
of Table~\ref{table:auto-sem} summarize the results for BLEU (n-gram
precision), METEOR (n-grams with synonym recall), CIDEr (weighted
n-gram cosine similarity), and NIST (weighted n-gram precision), 
where higher numbers indicate better overlap (shown with
the $\uparrow$). We note that while these measures are common 
for machine translation, they are not well-suited to this task, since
they are based on n-gram overlap which is not a constraint within the
model; we include them for comparative purposes.

From the table, we observe that across all metrics, we see a steady
increase as more information is added. Overall, the {{\sc +style}
  model has the highest scores for all metrics, i.e. {\sc +style}
  model outputs are most lexically similar to the references.\vspace{1em}
  
  \begin{table}[h!t]
\begin{footnotesize}
\begin{tabular}{p{0.01cm}p{1.2cm}p{0.1cm}p{0.8cm}p{0.8cm}p{0.8cm}p{1cm}}\toprule
 &  & & \sc Base          & \sc +Adj    & \sc +Sent  & \sc +Style    \\ \hline
1 & BLEU  & $\uparrow$ &	0.126  &	0.164  &	0.166  &\bf	0.173\\
2 & METEOR  & $\uparrow$ &	0.206  &	0.233  &	0.234  &\bf	0.235\\
3 & CIDEr  & $\uparrow$ &	1.300  &	1.686  &	1.692  &\bf	1.838\\
4 & NIST	  & $\uparrow$ & 3.840	 & 4.547	 & 4.477	 & \bf 5.537\\ \midrule
5 & Avg SER & $\downarrow$ &	\bf 0.053	& 0.063 &	0.064 &	0.090 \\ 
\bottomrule
\end{tabular}
\end{footnotesize}
\vspace{-.1in}
\centering \caption{\label{table:auto-sem} Automatic semantic evaluation (higher is better for all but SER).}
\end{table}

\noindent\textbf{Semantic Error Rate.} The {\it types of semantic errors} the
models make are more relevant than how well they conform to test
references. We calculate average Semantic Error Rate ($SER$), which is
a function of the number of semantic mistakes the model makes \cite{Wen15b,reedinlg18}. We find counts of two types of common mistakes: deletions, where the model fails to realize a value from the
input {\sc mr}, and repetitions, where the model repeats the same value
more than once.\footnote{{We note that other types of errors include insertions 
and substitutions, but we evaluate these through our human evaluation in 
Sec \ref{human-eval} since our large vocabulary size makes identifying them non-trivial.}} 
Thus, we compute SER per {\sc mr} as $SER = \frac{D + R}{N}$, where
$D$ and $R$ are the number of deletions and repetitions, and the $N$
is the number of tuples in the {\sc mr}, and average across the test outputs. 

Table \ref{table:auto-sem} presents the average SER rates for each
model, where lower rates mean fewer mistakes (indicated by
$\downarrow$). It is important to note here that we compute errors
over value and adjective slots only, since these are the ones that we
are able to identify lexically (we cannot identify whether an output
makes an error on sentiment in this way, so we measure that with a
human evaluation in Section~\ref{human-eval}). This means that the
{\sc base} outputs errors are computed over only value slots (since
they don't contain adjectives), and the rest of the errors are
computed over both value and adjective slots.

Amazingly, overall, Table \ref{table:auto-sem} results show the SER is
extremely low, even while achieving a large amount of stylistic
variation. Naturally, {\sc base}, with no access to style information,
has the best (lowest) SER. But we note that there is not a large
increase in SER as more information is added -- even for the most
difficult setting, {\sc +style}, the models make an error on less than
10\% of the slots in a given {\sc mr}, on average.

\subsection{Automatic Stylistic Evaluation}
We compute stylistic metrics to compare the model outputs, with results shown in Table \ref{table:auto-style}.\footnote{These measures can be compared to Table \ref{table:compare-corpora}, which includes similar statistics for the YelpNLG training data.} For {\bf vocab}, we find the number of unique words in all outputs for each model. We find the average {\bf sentence length} (SentLen) by counting the number of words, and find the total number of times an {\bf adjective} is used (Row 3) and average number of {adjectives} per reference for each model (Row 4). We compute Shannon text {\bf entropy} ($E$) as: $E = -\sum_{x\in V} \frac{f}{t} * log_2(\frac{f}{t})$, where $V$ is the vocab size in all outputs generated by the model, $f$ is the frequency of a term (in this case, a trigram), and $t$ counts the number of terms in all outputs. Finally, we count the instances of {\bf contrast} (e.g. "but" and "although"), and {\bf aggregation} (e.g. "both" and "also"). For all metrics, higher scores indicate more variability (indicated by $\uparrow$).

From the table, we see that overall the vocabulary is large, even
when compared to the training data for E2E and Laptop, as shown
in Table~\ref{table:compare-corpora}.  First, we see that the simplest, least constrained {\sc base} model has the largest vocabulary, since it has the most freedom in terms of word choice, while the model with the largest amount of supervision, {\sc +style}, has the smallest vocab, since we provide it with the most constraints on word
choice. For all other metrics, we see that the {\sc +style} model scores highest: these results are especially interesting when considering that {\sc +style} has the smallest vocab; even though word choice is constrained with richer style markup, {\sc +style} is more descriptive on average (more adjectives used), and has the highest entropy (more diverse word collocations). This is also very clear from the
significantly higher number of contrast and aggregation operations in the {\sc +style} outputs.\vspace{1em}

\begin{table}[t!]
\begin{footnotesize}
\begin{tabular}{p{0.01cm}p{1.1cm}p{0.1cm}p{0.9cm}p{0.9cm}p{0.9cm}p{0.9cm}}\toprule
 &  &  & \sc Base    & \sc +Adj    & \sc +Sent  & \sc +Style    \\ \hline
1 & Vocab & $\uparrow$ &	\bf 8,627	& 8,283	& 8,303	& 7,878 \\
2 & SentLen & $\uparrow$ &	11.27 &	11.45 &	11.30 	& \bf 13.91 \\ 
3 & \# Adjs& $\uparrow$& 24k & 26k & 26k & \bf 37k \\ 
4 & Adj/Ref& $\uparrow$& 0.82 & 0.90 & 0.89 & \bf1.26 \\ 
5 & Entropy & $\uparrow$ & 11.18 &	11.87 &	11.93 & \bf 11.94 \\\midrule
6 & Contrast & $\uparrow$ & 1,586 & 1,000 & 890 & \bf 2,769 \\ 
7 & Aggreg. & $\uparrow$ & 116 & 103 & 106 & \bf 1,178  \\
\bottomrule
\end{tabular}
\end{footnotesize}
\vspace{-.1in}
\centering \caption{\label{table:auto-style} Automatic stylistic evaluation metrics (higher is better). Paired t-test {\sc base} vs. {\sc +style} all $p < 0.05$.}
\end{table}
			
\noindent\textbf{Language Template Variations.} Since our
test set consists of 30k MRs, we are able to broadly characterize
and quantify the kinds of sentence constructions we get for each set
of model outputs.  To make generalized sentence templates, we delexicalize each
reference in the model outputs, i.e. we replace any food item with a
token {\sc [FOOD]}, any service item with {\sc [SERVICE]}, etc. Then,
we find the total number of unique templates each model produces, 
finding that each "more informed" model produces more unique
templates: {\sc base} produces 18k, {\sc +adj} produces 22k, {\sc
  +sent} produces 23k, and {\sc +style} produces 26k unique
templates. In other words, given the test set of 30k, {\sc +style}
produces a novel templated output for over 86\% of the input MRs.

While it is interesting to note that each "more informed" model
produces more unique templates, we also want to characterize {\it how
  frequently templates are reused}. Figure \ref{fig:template-reps}
shows the number of times each model repeats its top 20 most
frequently used templates. For example, the Rank 1 most frequently
used template for the {\sc base} model is {\it "I had the [FOOD]
  [FOOD]."}, and it is used 550 times (out of the 30k outputs). 
  For {\sc +style}, the Rank 1 most frequently
used template is {\it "I had the [FOOD] [FOOD] and it was
  delicious."}, and it is only used 130 times. The number of
repetitions decreases as the template rank moves from 1 to 20, 
and repetition count is always significantly lower for {\sc +style},
indicating more variation. Examples of frequent templates from the {\sc base} and {\sc +style} models are are shown in Appendix \ref{sec:app-templates}.\\

\begin{figure}[htb]
\begin{subfigure}[b]{0.9\columnwidth}
\centering
        \resizebox {\columnwidth} {!} {
\begin{tikzpicture}
\begin{axis}[
ymin=0.1,
legend style={at={(1,1)},anchor=north east,draw=none},
  xticklabel style = {rotate=30,anchor=east},
   enlargelimits = false,
      xtick={1, 5, 10, 15, 20},
   ytick={100,200,300,400,500},
   ylabel=Number of Repetitions,
   xlabel=Template Rank,
 ]
\addplot table [y=base,x=X]{repetition-dist.dat};
\addlegendentry{base}
\addplot table [y=+adj,x=X]{repetition-dist.dat};
\addlegendentry{+adj}
\addplot table [y=+sent,x=X]{repetition-dist.dat};
\addlegendentry{+sent}
\addplot table [y=+style,x=X]{repetition-dist.dat};
\addlegendentry{+style}
\end{axis}
\end{tikzpicture}
}
\end{subfigure}
\caption{Number of output template repetitions for the 20 most frequent templates ({\sc +Style} has the fewest repetitions, i.e. it is the most varied).}
    \label{fig:template-reps}
\end{figure}
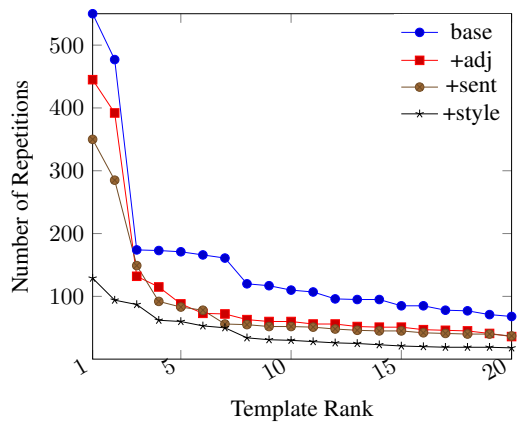

\noindent\textbf{Achieving Other Style Goals.} The {\sc +style} model is the only one with access to first-person, length, and exclamation markup, so we also measure its ability to hit these stylistic goals. The average sentence length for the {\sc +style} model for {\sc len=short} is 7.06 words, {\sc len=med} is 13.08, and {\sc len=long} is 22.74, closely matching the average lengths of the test references in those cases, i.e. 6.33, 11.05, and 19.03, respectively. The model correctly hits the target 99\% of the time for first person (it is asked to produce this for 15k of the 30k test instances), and 100\% of the time for exclamation (2k instances require exclamation).

\subsection{Human Quality Evaluation} 
\label{human-eval}
We evaluate output quality using human annotators on Mechanical
Turk. As in our corpus quality evaluation from Section \ref{sec:corpus-quality}, we 
randomly sample 200 {\sc mr}s from the test
set, along with the corresponding outputs for each of the 4
models, and ask 5 annotators to rate each output on a 1-5 Likert scale
for {\bf content}, {\bf fluency}, and {\bf sentiment} (1 for very
negative, 5 for very positive\footnote{As in Sec \ref{sec:corpus-quality}, we scale
the sentiment scores into 3 bins to match our Yelp review sentiment.}). Table
\ref{table:human-qual} shows the average scores by criteria and model.\footnote{The average correlation between each
  annotator's ratings and the average rating for each item is 0.73.}

For content and fluency, all average ratings are very
high, above 4.3 (out of 5). The differences between models are small, but
it is interesting to note that the {\sc base} and {\sc +style} models
are almost tied on fluency (although {\sc base} outputs may
appear more fluent due to their comparably shorter length). In the
case of sentiment error, the largest error is 0.75 {(out of 3)}, with the smallest
sentiment error (0.56) achieved by the {\sc +style} model. Examination of
the outputs reveals that the most common sentiment error is producing
a neutral sentence when negative sentiment is specified. This may be
due to the lower frequency of negative sentiment in the corpus as well
as noise in automatic sentiment annotation.

\begin{table}[!h]
\begin{footnotesize}
\begin{tabular}{p{1.8cm}p{0.1cm}p{0.7cm}p{0.7cm}p{0.7cm}p{0.9cm}}\toprule
 & & \sc Base          & \sc +Adj    & \sc +Sent  & \sc +Style    \\ \hline
Content & $\uparrow$ & 4.35*	&	 4.53&	4.51 & 4.49 \\
Fluency & $\uparrow$&  4.43	&	4.36&	4.37 & 4.41 \\
Sentiment Err &  $\downarrow$&0.75* &	0.71*&	 0.67* &  0.56 \\
\bottomrule
\end{tabular}
\end{footnotesize}
\vspace{-.1in}
\centering \caption{\label{table:human-qual} Human quality evaluation (higher is better for content and fluency, lower is better for sentiment error). Paired t-test for each model vs.{\sc +style}, * is $p < 0.05$. }
\end{table}
\vspace{-.1in}

\section{Related Work}
\label{sec:related}

Recent efforts on data acquisition for {\sc nnlg} has relied almost exclusively on crowdsourcing. Novikova et al. \shortcite{novikova2017e2e} used pictorial representations of restaurant {\sc mr}s to elicit 50k varied restaurant descriptions through crowdsourcing. Wen et al. \shortcite{Wen15b,Wenetal15} also create datasets for the restaurant (5k), hotel (5k), laptop (13k), and TV (7k) domains by asking Turkers to write {\sc nl} realizations for different combinations of input dialog acts in the {\sc mr}. Work on the {\sc WebNLG} challenge has also focused on using existing structured data, such as DBPedia, as input into an {\sc nlg} \cite{nlg_microplanning}, where matching {\sc nl} utterances are also crowdsourced. {Other recent work on collecting datasets for dialog modeling also use large-scale crowdsourcing \cite{budzianowski-etal-2018-multiwoz}.} 



Here, we completely avoid having to crowdsource any data by
working in reverse: we begin with naturally occurring user reviews,
and automatically construct {\sc mr}s from them. This allows
us to create a novel dataset {\sc YelpNLG}, the 
largest existing {\sc nlg} dataset, with 300k 
parallel {\sc mr} to sentence pairs with rich information on attribute,
value, description, and mention order, in addition to a set of sentence-level style information, including sentiment, length, and pronouns.


{ In terms of control mechanisms, very recent work in {\sc nnlg} has begun to explore using an explicit sentence planning stage and  hierarchical structures \cite{moryossef-separating,balakrishnan-constrained}. In our own work, we show how we are able to control various aspects of style with simple supervision within the input {\sc mr}, without requiring a dedicated sentence planner, and in line with the end-to-end neural generation paradigm. }

Previous work has primarily attempted to {\it individually} control aspects of content
preservation and style attributes such as formality and verb tense,
sentiment \shortcite{shen2017style}, and personality in different
domains such as news and product reviews \cite{fu2018style}, movie
reviews \cite{Ficler17,hu2017toward}, restaurant descriptions
\cite{oraby2018stylepersonage}, and customer care dialogs
\cite{herzig2017neural}. To our knowledge, our work is the very first
to generate realizations that both express particular semantics {\it
  and} exhibit a particular descriptive or lexical style and
sentiment. It is also the first work to our knowledge that controls
lexical choice in neural generation, a long standing interest of the
NLG community
\cite{BarzilayLee02,Elhadad92,radev98learning,MoserMoore95,Hirschberg08}.

\section{Conclusions}
\label{sec:conc}

This paper presents the YelpNLG corpus, a set of 300,000
parallel sentences and {\sc mr} pairs generated by sampling freely
available review sentences that contain attributes of interest, and
automatically constructing {\sc mr}s for them. The dataset is unique in its huge range of
stylistic variation and language richness, particularly compared to
existing parallel corpora for {\sc nlg}. We train different
models with varying levels of information related to attributes,
adjective dependencies, sentiment, and style information, and
present a rigorous set of evaluations to
quantify the effect of the style markup on the ability of the models
to achieve multiple style goals.
 
For future work, {we plan on exploring other models for {\sc nlg}}, and on 
providing models with a more detailed
input representation in order to help preserve more dependency
information, as well as to encode more information on syntactic
structures we want to realize in the output. {We are also interested in including richer, more semantically-grounded information in our {\sc mr}s, for example using Abstract Meaning Representations ({\sc amr}s) \cite{Dorr:1998:THE:648179.749214,Banarescu13abstractmeaning,flanigan-etal-2014-discriminative}.} 
Finally, we are interested in
reproducing our corpus generation method on various other domains to
allow for the creation of numerous useful datasets for the {\sc nlg}
community.

\bibliography{nlg,phd,nl}
\bibliographystyle{acl_natbib}


\appendix
\newpage
{\bf \noindent \large Appendix}

\section{Model Configurations}
\label{sec:app-model}
Here we describe final model configurations 
for the most complex model, {\sc +style}, after experimenting with 
different parameter settings. The encoder and decoder are each three layer LSTMs with 600 units.  
We use Dropout \cite{Srivastava_Hinton_Krizhevsky_Sutskever_Salakhutdinov_2014} 
of 0.3 between RNN layers. Model parameters are initialized using 
Glorot initialization \cite{Glorot_Bengio_2010} and are optimized using stochastic gradient descent with mini-batches of size 64. We use a learning rate of 1.0 with a decay rate of 0.5 that gets applied after each training epoch starting with the fifth epoch. Gradients are clipped when the absolute value is greater than 5. 
We tune model hyper-parameters on a development dataset
and select the model of lowest perplexity to evaluate on the test dataset.
Beam search with three beams is used during inference.
{\sc mr}s are represented using 300 dimensional embeddings. The target side word embeddings are initialized using pre-trained Glove word vectors \cite{Glove_Pennington_Socher_Manning_2014} which get updated during training.   Models are trained using lowercased reference texts.

%

\newpage
\section{Repeated Templates from {\sc base} and {\sc +style}}
\label{sec:app-templates}

Table \ref{table:templates} shows the top 10 most repeated templates for the {\sc base} and {\sc +style} models. Note that "\# Reps" indicates the number of times the template is repeated in the test set of 30k instances; the largest number of reps is only 550 for the most frequent {\sc base} model template, only 129 for {\sc +style}, meaning that the models mostly generate novel outputs for each test instance.

\begin{table}[h!]
\begin{small}
\begin{tabular}
{@{} p{0.4in}p{2.4in} @{}} \toprule
\# Reps & {\sc base} Templates \\ \hline
550 & {\it i had the [FOOD] [FOOD].}\\
477 & {\it i had the [FOOD] and [FOOD].}\\
174 & {\it i had the [FOOD] [FOOD] [FOOD].}\\
173 & {\it the [FOOD] [FOOD] was good.}\\
171 & {\it the [FOOD] and [FOOD] were good.}\\
166 & {\it the [FOOD] was tender and the [FOOD] was delicious.}\\
161 & {\it i had the [FOOD] fried [FOOD].}\\
120 & {\it the [FOOD] [FOOD] was very good.}\\
117 & {\it the [FOOD] was good but the [FOOD] was a little dry.}\\ \midrule

& {\sc +style} Templates \\ \hline
129 & {\it i had the [FOOD] [FOOD] and it was delicious.}\\
94 & {\it had the [FOOD] and [FOOD] [FOOD] plate.}\\
87 & {\it the [FOOD] and [FOOD] were cooked to perfection.}\\
62 & {\it i had the [FOOD] [FOOD] and it was good.}\\
60 & {\it i had the [FOOD] [FOOD].}\\
53 & {\it i had the [FOOD] and my husband had the [FOOD].}\\
50 & {\it i had the [FOOD] and [FOOD] and it was delicious.}\\
34 & {\it the [FOOD] and [FOOD] skewers were the only things that were good.}\\
31 & {\it i had the [FOOD] [FOOD] [FOOD] and it was delicious.}\\

\bottomrule
\end{tabular}
 \caption{Sample of 10 "most repeated" templates from {\sc base} and {\sc +style}.\label{table:templates}}
\end{small}
\end{table}

 \section{Sample Model Outputs for {\sc +style}}
{Table \ref{table:model-op} shows examples outputs from the {\sc +style} model, with specific examples of style through different forms of personal pronoun use, contrast, aggregation, and hyperbole in Tables \ref{table:model-op-pronouns}-\ref{table:model-op-hyperbole}.}

\label{sec:app-outputs}
\vspace{-2cm}
\begin{table*}[hbt!]
\begin{small}
\begin{tabular}
{@{} p{0.001in}p{6in} @{}} \toprule
1 & {\it (attr=}food,  {\it val=}meat,  {\it adj=}chewy,  {\it mention=}1), ({\it attr=}food,  {\it val=}sauce,  {\it adj=}no-adj,  {\it mention=}1), \newline +[{\it sentiment=}negative,  {\it len=}medium,  {\it first-person=}false,  {\it exclamation=}false]\\ 
& {\bf The meat was chewy and the sauce had no taste.}   \\
\midrule

2 & {\it (attr=}food,  {\it val=}artichokes,  {\it adj=}no-adj,  {\it mention=}1), 
 {\it (attr=}food,  {\it val=}beef-carpaccio,  {\it adj=}no-adj,  {\it mention=}1), 
 +[{\it sentiment=}positive,  {\it len=}long,  {\it first-person=}true,  {\it exclamation=}false]\\ 
& {\bf We started with the artichokes and beef carpaccio , which were the highlights of the meal .}   \\\midrule

3 & {\it (attr=}staff,  {\it val=}waitress,  {\it adj=}no-adj,  {\it mention=}1), 
 {\it (attr=}food,  {\it val=}meat-tips,  {\it adj=}no-adj,  {\it mention=}1), 
 {\it (attr=}food,  {\it val=}ribs,  {\it adj=}no-adj,  {\it mention=}1), 
 +[{\it sentiment=}neutral,  {\it len=}long,  {\it first-person=}true,  {\it exclamation=}false]\\ 
& {\bf The waitress came back and told us that they were out of the chicken meat tips and ribs .}   \\ \midrule

4 & {\it (attr=}food,  {\it val=}chicken-lollipops,  {\it adj=}good,  {\it mention=}1), 
 {\it (attr=}food,  {\it val=}ambiance,  {\it adj=}nice,  {\it mention=}1), 
 +[{\it sentiment=}positive,  {\it len=}medium,  {\it first-person=}false,  {\it exclamation=}false]\\ 
& {\bf The chicken lollipops were really good , nice ambience .}   \\\midrule

5 & {\it (attr=}food,  {\it val=}meat,  {\it adj=}no-adj,  {\it mention=}1), 
 {\it (attr=}food,  {\it val=}sausage,  {\it adj=}no-adj,  {\it mention=}1), 
 {\it (attr=}food,  {\it val=}deli-meats,  {\it adj=}no-adj,  {\it mention=}1), 
 {\it (attr=}food,  {\it val=}cheeses,  {\it adj=}no-adj,  {\it mention=}1), 
 {\it (attr=}price,  {\it val=}prices,  {\it adj=}good,  {\it mention=}1), 
 +[{\it sentiment=}positive,  {\it len=}medium,  {\it first-person=}false,  {\it exclamation=}false]\\ 
& {\bf Geat selection of meat , sausage , deli meats , cheeses , and good prices .}   \\\midrule

6 & {\it (attr=}food,  {\it val=}beef-chili,  {\it adj=}amazing,  {\it mention=}1), 
 {\it (attr=}food,  {\it val=}onion,  {\it adj=}carmalized,  {\it mention=}1), 
 +[{\it sentiment=}positive,  {\it len=}long,  {\it first-person=}true,  {\it exclamation=}false]\\ 
& {\bf The beef chili was amazing , and i loved the caramelized onions that came with it .}   \\\midrule


7 & {\it (attr=}food,  {\it val=}eggs,  {\it adj=}runny,  {\it mention=}1), 
 {\it (attr=}food,  {\it val=}crab-legs,  {\it adj=}open,  {\it mention=}1), 
 +[{\it sentiment=}neutral,  {\it len=}long,  {\it first-person=}true,  {\it exclamation=}false]\\ 
& {\bf The eggs were runny , and the open faced crab legs were a little too much for my taste .}   \\\midrule

8 & {\it (attr=}food,  {\it val=}chicken-salad,  {\it adj=}grilled,  {\it mention=}1), 
 {\it (attr=}food,  {\it val=}chicken,  {\it adj=}no-adj,  {\it mention=}1), 
 {\it (attr=}food,  {\it val=}spices,  {\it adj=}right,  {\it mention=}1), 
 {\it (attr=}food,  {\it val=}salad,  {\it adj=}fresh,  {\it mention=}1), 
 +[{\it sentiment=}positive,  {\it len=}long,  {\it first-person=}true,  {\it exclamation=}false]\\ 
& {\bf I had the grilled chicken salad , the chicken was tender and the spices and fresh salad were just right .}   \\
\bottomrule
\end{tabular}
 \caption{Sample test outputs from Model {\sc +style}.\label{table:model-op}}
\end{small}
\end{table*}

\begin{table*}[h!]
\begin{small}
\begin{tabular}
{@{} p{0.001in}p{5.8in} @{}} \toprule
1 & {\it (attr=}food,  {\it val=}fish-meat,  {\it adj=}no-adj,  {\it mention=}1), 
 {\it (attr=}food,  {\it val=}horse-radish-sauce,  {\it adj=}no-adj,  {\it mention=}1), 
 +[{\it sentiment=}positive,  {\it len=}long,  {\it first-person=}true,  {\it exclamation=}false]\\
& {\bf I had the fish meat and it was very good, and my husband had the chicken horse-radish-sauce which he loved.}   \\\midrule

2 & {\it (attr=}food,  {\it val=}beef,  {\it adj=}no-adj,  {\it mention=}1), 
 {\it (attr=}restaurant,  {\it val=}restaurant,  {\it adj=}nice,  {\it mention=}1), 
 {\it (attr=}staff,  {\it val=}waiter,  {\it adj=}friendly,  {\it mention=}1), 
 +[{\it sentiment=}positive,  {\it len=}long,  {\it first-person=}true,  {\it exclamation=}false]\\
& {\bf The beef was tender, the restaurant was nice, and the waiter was friendly and helpful to us.}   \\\midrule

3 & {\it (attr=}food,  {\it val=}lobster,  {\it adj=}no-adj,  {\it mention=}1), 
 {\it (attr=}food,  {\it val=}crab-legs,  {\it adj=}no-adj,  {\it mention=}1), 
 {\it (attr=}food,  {\it val=}mussels,  {\it adj=}no-adj,  {\it mention=}1), 
 {\it (attr=}food,  {\it val=}clams,  {\it adj=}no-adj,  {\it mention=}1), 
 +[{\it sentiment=}positive,  {\it len=}medium,  {\it first-person=}true,  {\it exclamation=}false]\\
& {\bf We had lobster, crab legs, mussels and clams.} \\ \midrule

4 & {\it (attr=}food,  {\it val=}crab-soup,  {\it adj=}no-adj,  {\it mention=}1), 
 +[{\it sentiment=}negative,  {\it len=}short,  {\it first-person=}false,  {\it exclamation=}false]\\ 
& {\bf She had the crab soup.}   \\ \midrule

5 & {\it (attr=}staff,  {\it val=}host,  {\it adj=}no-adj,  {\it mention=}1), 
 {\it (attr=}food,  {\it val=}steak,  {\it adj=}no-adj,  {\it mention=}1), 
 {\it (attr=}food,  {\it val=}lobster,  {\it adj=}no-adj,  {\it mention=}1), 
 +[{\it sentiment=}positive,  {\it len=}long,  {\it first-person=}false,  {\it exclamation=}false]\\
& {\bf The host came out with the steak and lobster, and he said it was very good .}
\\ \bottomrule
\end{tabular}
 \caption{Examples of different pronouns from Model {\sc +style}.\label{table:model-op-pronouns}}
\end{small}
\end{table*}

\begin{table*}[h!]
\begin{small}
\begin{tabular}
{@{} p{0.001in}p{5.8in} @ {}} \toprule
1 & {\it (attr=}food,  {\it val=}kids-chicken-fingers,  {\it adj=}no-adj,  {\it mention=}1), 
 {\it (attr=}food,  {\it val=}chicken,  {\it adj=}actual,  {\it mention=}1), 
 {\it (attr=}food,  {\it val=}chicken,  {\it adj=}little,  {\it mention=}2), 
 +[{\it sentiment=}positive,  {\it len=}long,  {\it first-person=}false,  {\it exclamation=}false]\\ 
& {\bf The kids chicken fingers are made with actual chicken, but the chicken is a little on the dry side.}   \\ \midrule

2 & {\it (attr=}food,  {\it val=}nachos,  {\it adj=}no-adj,  {\it mention=}1), 
 {\it (attr=}food,  {\it val=}chicken,  {\it adj=}no-adj,  {\it mention=}1), 
 +[{\it sentiment=}negative,  {\it len=}long,  {\it first-person=}true,  {\it exclamation=}false]\\
& {\bf I ordered the nachos with chicken, and they were pretty good, but nothing to write home about.}   \\ \midrule
%

3 & {\it (attr=}food,  {\it val=}chicken-tenders,  {\it adj=}no-adj,  {\it mention=}1), 
 {\it (attr=}food,  {\it val=}chicken-nuggets,  {\it adj=}no-adj,  {\it mention=}1), 
 +[{\it sentiment=}neutral,  {\it len=}long,  {\it first-person=}true,  {\it exclamation=}false]\\
& {\bf The chicken tenders and chicken nuggets were the only things that were good, but nothing special.} \\ \midrule

4 & {\it (attr=}food,  {\it val=}rice,  {\it adj=}good,  {\it mention=}1), 
 {\it (attr=}food,  {\it val=}meat,  {\it adj=}no-adj,  {\it mention=}1), 
 +[{\it sentiment=}neutral,  {\it len=}long,  {\it first-person=}true,  {\it exclamation=}false]\\
& {\bf The rice was good, but i wish there was more meat in the dish.}
\\ \bottomrule
\end{tabular}
 \caption{Examples of contrast from Model {\sc +style}.\label{table:model-op-contrast}}
\end{small}
\end{table*}

\begin{table*}[h!]
\begin{small}
\begin{tabular}
{@{} p{0.001in}p{5.8in} @{}} \toprule
1 & {\it (attr=}food,  {\it val=}meat,  {\it adj=}no-adj,  {\it mention=}1), 
 {\it (attr=}food,  {\it val=}sausage,  {\it adj=}no-adj,  {\it mention=}1), 
 {\it (attr=}food,  {\it val=}deli-meats,  {\it adj=}no-adj,  {\it mention=}1), 
 {\it (attr=}food,  {\it val=}cheeses,  {\it adj=}no-adj,  {\it mention=}1), 
 {\it (attr=}price,  {\it val=}prices,  {\it adj=}good,  {\it mention=}1), 
 +[{\it sentiment=}positive,  {\it len=}medium,  {\it first-person=}false,  {\it exclamation=}false]\\ 
& {\bf Great selection of meat, sausage, deli meats, cheeses, and good prices.}    \\ \midrule
2 & {\it (attr=}food,  {\it val=}tofu,  {\it adj=}fried,  {\it mention=}1), 
 {\it (attr=}food,  {\it val=}lemongrass-chicken,  {\it adj=}aforementioned,  {\it mention=}1), 
 +[{\it sentiment=}neutral,  {\it len=}long,  {\it first-person=}true,  {\it exclamation=}false]\\
& {\bf I had the fried tofu and my husband had the lemongrass chicken, both of which were very good.}    \\ \midrule
3 & {\it (attr=}food,  {\it val=}burgers,  {\it adj=}different,  {\it mention=}1), 
 {\it (attr=}food,  {\it val=}chicken-club,  {\it adj=}grilled,  {\it mention=}1), 
 +[{\it sentiment=}positive,  {\it len=}long,  {\it first-person=}true,  {\it exclamation=}false]\\
& {\bf We ordered two different burgers and a grilled chicken club, both of which were delicious.}    \\ \midrule
4 & {\it (attr=}food, {\it val=}octopus,  {\it adj=}no-adj,  {\it mention=}1),
{\it (attr=}food, {\it val=}salmon,  {\it adj=}no-adj, {\it mention=}1),
{\it (attr=}food, {\it val=}tuna,  {\it adj=}no-adj,  {\it mention=}1),
{\it (attr=}food, {\it val=}crab,  {\it adj=}no-adj,  {\it mention=}1),
{\it (attr=}food, {\it val=}squid,  {\it adj=}no-adj,  {\it mention=}1),
{\it (attr=}food, {\it val=}shrimp,  {\it adj=}no-adj,  {\it mention=}1),
+[{\it sentiment=}positive, {\it len=}long, {\it first-person=}false,  {\it exclamation=}true]\\
& {\bf Octopus, salmon, tuna, crab, squid, shrimp, etc... all of it was delicious !}    
\\ \bottomrule
\end{tabular}
 \caption{Examples of aggregation from Model {\sc +style}.\label{table:model-op-aggreg}}
\end{small}
\end{table*}

\begin{table*}[h!]
\begin{small}
\begin{tabular}
{@{} p{0.001in}p{5.8in} @{}} \toprule
   
1 & {\it (attr=}food,  {\it val=}meat,  {\it adj=}spectacular,  {\it mention=}1), 
 {\it (attr=}food,  {\it val=}sauces,  {\it adj=}no-adj,  {\it mention=}1), 
 +[{\it sentiment=}positive,  {\it len=}medium,  {\it first-person=}false,  {\it exclamation=}false]\\ 
& {\bf The meat was spectacular and the sauces were to die for. }   \\ \midrule

2 & {\it (attr=}food,  {\it val=}maine-lobster,  {\it adj=}heavenly,  {\it mention=}1), 
 {\it (attr=}food,  {\it val=}crab-bisque,  {\it adj=}no-adj,  {\it mention=}1), 
 +[{\it sentiment=}positive,  {\it len=}long,  {\it first-person=}false,  {\it exclamation=}false]\\ 
& {\bf The lobster claw was heavenly, and the crab bisque was a nice touch, but not overpowering.}   \\ \midrule

3 & {\it (attr=}food,  {\it val=}meat-sauce-spaghetti,  {\it adj=}no-adj,  {\it mention=}1),
 {\it (attr=}food,  {\it val=}milk-tea,  {\it adj=}cold,  {\it mention=}1), 
 +[{\it sentiment=}positive,  {\it len=}long,  {\it first-person=}true,  {\it exclamation=}false]\\ 
& {\bf I had the chicken meat sauce spaghetti and it was very good and the cold milk tea was the best i have ever had.}   \\ \midrule

4 & {\it (attr=}food,  {\it val=}seafood,  {\it adj=}fresh,  {\it mention=}1), 
 {\it (attr=}food,  {\it val=}chicken,  {\it adj=}fried,  {\it mention=}1), 
 {\it (attr=}food,  {\it val=}bread-pudding,  {\it adj=}phenomenal,  {\it mention=}1), 
 +[{\it sentiment=}positive,  {\it len=}long,  {\it first-person=}false,  {\it exclamation=}false]\\ 
& {\bf The seafood was fresh, the fried chicken was great, and the bread pudding was phenomenal.}  
\\ \bottomrule
\end{tabular}
 \caption{Examples of hyperbole from Model {\sc +style}.\label{table:model-op-hyperbole}}
\end{small}
\end{table*}


\end{document}